\documentclass[10pt, conference, letterpaper]{IEEEtran}
\IEEEoverridecommandlockouts
\usepackage{cite}
\usepackage{amsmath,amssymb,amsfonts}
\usepackage{algorithmic}
\usepackage{graphicx}
\usepackage{textcomp}
\usepackage{xcolor}
\usepackage[ruled,vlined,linesnumbered]{algorithm2e}
\def\BibTeX{{\rm B\kern-.05em{\sc i\kern-.025em b}\kern-.08em
    T\kern-.1667em\lower.7ex\hbox{E}\kern-.125emX}}
\begin{document}

\title{Bypassing Krum: Selection-Aware Backdoor Attacks in Federated Learning%
\thanks{This work was accepted and presented at the 2026 International Conference on Intelligent Multimedia, Networking, and Security (IMNS 2026).}}

\author{\IEEEauthorblockN{Srinivasan Subramanian}
\IEEEauthorblockA{\textit{Department of Computer Science} \\
\textit{Kennesaw State University}\\
Marietta, USA \\
ssubram7@students.kennesaw.edu}
\and
\IEEEauthorblockN{Md.~Abdullah~Al~Hafiz~Khan}
\IEEEauthorblockA{\textit{Department of Computer Science} \\
\textit{Kennesaw State University}\\
Marietta, USA \\
mkhan74@kennesaw.edu}
\and
\IEEEauthorblockN{Kazi Aminul Islam}
\IEEEauthorblockA{\textit{Department of Computer Science} \\
\textit{Kennesaw State University}\\
Marietta, USA \\
kislam4@kennesaw.edu}
}

\maketitle

\begin{abstract}
Robust aggregation methods are widely used in federated learning to mitigate the impact of adversarial client behavior. Distance-based aggregation rules, such as Krum and Multi-Krum, select updates that are closest to the majority under the assumption that benign updates form a compact cluster. However, these methods rely on geometric properties that can be exploited by adaptive adversaries. We introduce the Krum-Proxy attack, a selection-aware backdoor injection strategy that consistently bypasses Byzantine-robust aggregation. Rather than relying on naive scaling or constraining, our method actively optimizes malicious updates to infiltrate the dense core of the benign distribution. The proposed method constructs adversarial updates that are not only similar to benign updates but are also optimized to lie in regions of the update space that are favored during aggregation. This is achieved through a two-stage optimization procedure that separates task-specific attack objectives from geometry-aware refinement, using a nearest-neighbor proxy, stochastic reference modeling, and anchor-guided alignment. To maintain stealth, we introduce a projection mechanism that constrains adversarial updates within realistic norm and variance bounds. Experiments on standard federated learning benchmarks show that Krum-Proxy achieves higher attack success while preserving clean accuracy, highlighting the vulnerability of distance-based aggregation to selection-aware adversaries.
\end{abstract}


\section{Introduction}

Federated Learning (FL) enables collaborative model training across decentralized data sources while preserving data privacy \cite{b11}. However, the server relies on client-provided updates without direct access to local data, making it vulnerable to adversarial manipulation.

Model poisoning and backdoor attacks are among the most critical threats in FL, where malicious clients inject carefully crafted updates to manipulate the global model while maintaining a benign appearance \cite{b3}. These attacks can embed targeted behaviors that are triggered only under specific inputs, making them difficult to detect. To mitigate such threats, Byzantine-resilient methods such as Krum and Multi-Krum have been proposed \cite{b1}. These methods select updates closest to the majority based on pairwise distances, assuming benign updates form a compact cluster.

Despite their theoretical guarantees, Krum-based defenses are vulnerable to adaptive adversaries. Prior work has shown that malicious updates can be crafted to mimic the statistical properties of benign updates, allowing them to bypass distance-based filtering \cite{b4, b5}. Since Krum relies on a single scalar distance over the entire model update, it fails to capture structured deviations across layers \cite{b2, b4}, allowing adversarial signals to remain hidden.

Our work differs from previous Krum-aware and optimization-based attacks \cite{b5, b15, b16} by reducing the attacker’s dependence on real-time benign updates while still explicitly targeting the geometry used by Krum-style aggregation. The attack uses attacker-local stochastic references to approximate benign update structure, optimizes a differentiable proxy of Krum’s neighborhood score, and separates backdoor injection from geometry shaping through a two-stage schedule. Our contributions are as follows.

\begin{itemize}

\item \textbf{Selection-aware attack formulation for Krum-based aggregation:}
We propose \emph{Krum-Proxy}, a targeted backdoor attack that explicitly optimizes for the selection rule of Krum, rather than relying only on norm matching, scaling, or directional constraints. The attack uses a differentiable approximation of Krum scoring constructed from attacker-local reference updates.

\item \textbf{Stochastic Reference Construction:}
We introduce a stochastic reference modeling strategy that approximates benign update geometry using attacker-trained clean deltas with varied training conditions. This captures benign variability and enables local estimation of neighborhood structure for Krum scoring without access to real client updates.

\item \textbf{Two-Stage Optimization:}
We develop a two-stage optimization procedure that separates (i) backdoor injection with cross-entropy loss from (ii) geometry shaping through norm matching, proxy distance, and anchor losses with a reduced learning rate. This decoupling prevents competing objectives from interfering with early backdoor convergence and enables stable selection-aware refinement in the second stage.

\item \textbf{Anchor-Guided Alignment and Projection:}
We introduce an anchor-guided alignment and projection mechanism that steers adversarial updates toward high-density benign regions while preserving the backdoor signal through controlled norm and direction constraints.

\item \textbf{Comprehensive Empirical Validation:}
Experiments demonstrate that Krum-Proxy achieves higher and more stable attack success than scaling-based and constrain-and-scale baselines while maintaining competitive clean accuracy. We also report malicious selection and inclusion rates, directly validating that the attack improves aggregation-level evasion.

\end{itemize}

\section{Related Work}

The standard FL aggregation method, FedAvg, is efficient but highly vulnerable to adversarial manipulation, as malicious clients can arbitrarily influence the global model \cite{b3}. To address this, Byzantine-robust aggregation rules such as Krum and Multi-Krum were proposed, which select updates closest to the majority based on pairwise distances \cite{b1}. 

However, prior work has shown that distance-based aggregation degrades in heterogeneous and high-dimensional settings, weakening its robustness guarantees in practice \cite{b2,b9,b10}. Alternative aggregation methods attempt to mitigate these issues, but still rely on assumptions that adversaries can exploit \cite{b13,b14}.

Model poisoning attacks exploit this vulnerability by directly manipulating client updates during training. Early work demonstrated that adversaries can induce targeted misbehavior or degrade model performance even with limited control \cite{b7, b15}. Backdoor attacks further increase stealth by embedding malicious behaviors that are triggered only with specific input. Bagdasaryan et al. \cite{b3} introduced the Constrain-and-Scale attack, which jointly optimizes attack success and stealth through norm and direction constraints. Subsequent work showed that such attacks can persist across rounds and remain effective with partial participation \cite{b8, b17}. Distributed attacks, such as DBA \cite{b6}, further improve robustness by splitting the attack across multiple clients.

Building on this, Baruch et al. \cite{b4} demonstrated that small but carefully aligned perturbations can bypass defenses by exploiting benign variance. Fang et al. \cite{b5} proposed local poisoning attacks that explicitly target aggregation rules such as Krum. Shejwalkar and Houmansadr \cite{b16} further formulated poisoning as an optimization problem, allowing attackers to systematically adapt to defense mechanisms. 

Krum-Proxy differs from these aggregation-aware attacks in both access assumptions and optimization structure. Fang et al. \cite{b5} directly target Byzantine-robust aggregation, but rely on knowledge of the update population when optimizing against the aggregation rule. Shejwalkar et al. \cite{b16} formulate defense-aware poisoning through MinMax and MinSum objectives, but these objectives impose global distance constraints rather than a local proxy of Krum's nearest-neighbor selection. In contrast, Krum-Proxy constructs attacker-local stochastic references, derives a differentiable proxy Krum score from their inner neighborhood, and applies it after a separate backdoor injection stage.

\section{Proposed Method: Selection-Aware Two-Stage Krum-Proxy Attack}

We propose a geometry-aware model poisoning attack that explicitly targets the selection mechanism of Krum and Multi-Krum. The attack consists of three components: (i) stochastic reference construction, (ii) two-stage adversarial optimization, and (iii) projection-based stealth enforcement.

\subsection{Threat Model}

We assume a gray-box adversary controlling $f$ out of $N$ clients ($f < N/2$). The attacker knows that the server uses Krum or Multi-Krum, but does not observe benign client updates or server-side scores at attack time. The attacker has labeled local data drawn from a distribution comparable to benign clients, with target-task label coverage sufficient for backdoor training and clean reference construction. In our implementation, the attacker reconstructs $R=6$ stochastic references per attack round and uses $E_1=E_2=1$ additional local epochs for the two attack stages. The goal is to induce a target label on the triggered inputs while maintaining clean accuracy.

\begin{algorithm}[!b]
\DontPrintSemicolon
\SetKwInOut{Input}{Input}
\SetKwInOut{Output}{Output}
\SetKwInOut{Param}{Param}

\Input{Global model $w_t$; local data $\mathcal{D}$; reference set 
       $\{\Delta_r\}_{r=1}^R$; local benign delta $\Delta_{\text{loc}}$}
\Param{$\eta$, $E_1$, $E_2$, $k$, 
       $(\lambda_l, \lambda_n, \lambda_k, \lambda_a)$, 
       clip $\tau{=}5.0$}
\Output{Adversarial model $w_{\text{adv}}$}
\BlankLine

\tcp{Anchor construction}
$s_r \leftarrow \frac{1}{k}\!\sum_{j \in \mathcal{N}_k(r)} 
   \|\Delta_r - \Delta_j\|^2,\ \forall r$\;
$(1), (2) \leftarrow \arg\text{sort}_r\, s_r$ \quad (top-two lowest)\;
$\Delta_{\text{anchor}} \leftarrow 0.85\,\Delta_{(1)} + 0.15\,\Delta_{(2)}$\;
\BlankLine

\tcp{Stage 1: Backdoor injection}
$w \leftarrow w_t$\;
\For{$e = 1$ \KwTo $E_1$}{
  \ForEach{mixed batch $(x, y)$ of clean + trigger-poisoned samples}{
    $g \leftarrow \nabla_w \mathcal{L}_{\text{CE}}(f_w(x), y)$\;
    clip $\|g\| \leq \tau$;\quad 
    $w \leftarrow \text{SGD}(w, g, \eta, \mu{=}0.9)$\;
  }
}
\BlankLine

\tcp{Stage 2: Geometry shaping}
\For{$e = 1$ \KwTo $E_2$}{
  \ForEach{mixed batch $(x, y)$}{
    $\Delta_{\text{adv}} \leftarrow w - w_t$\;
    $d_j \leftarrow \|\Delta_{\text{adv}} - \Delta_j\|^2$\;
    $\mathcal{N} \leftarrow \text{top-}k\,(\{d_j\},\, \text{smallest})$\;
    $\mathcal{L}_{\text{krum}} \leftarrow \frac{1}{\lfloor k/2 \rfloor} 
       \sum_{j \in \mathcal{N}[:\lfloor k/2 \rfloor]} d_j$\;
    $\mathcal{L}_{\text{norm}} \leftarrow 
       (\|\Delta_{\text{adv}}\| - \|\Delta_{\text{loc}}\|)^2$\;
    $\mathcal{L}_{\text{anchor}} \leftarrow 
       \|\Delta_{\text{adv}} - \Delta_{\text{anchor}}\|^2$\;
    $\mathcal{L} \leftarrow \lambda_l \mathcal{L}_{\text{CE}}
       + \lambda_n \mathcal{L}_{\text{norm}}
       + \lambda_k \mathcal{L}_{\text{krum}}
       + \lambda_a \mathcal{L}_{\text{anchor}}$\;
    $g \leftarrow \nabla_w \mathcal{L}$;\quad clip $\|g\| \leq \tau$\;
    $w \leftarrow \text{SGD}(w, g, 0.5\eta, \mu{=}0.9)$\;
  }
}
\BlankLine

\tcp{Projection}
$\Delta_{\text{adv}} \leftarrow w - w_t$\;
$\sigma_{\text{ref}} \leftarrow \text{std}_r\,
   \|\Delta_r - \bar{\Delta}_{\text{ref}}\|$\;
\If{$\|\Delta_{\text{adv}} - \Delta_{(1)}\| > \sigma_{\text{ref}}$}{
  $\alpha \leftarrow \mathrm{clip}\!\left(
    \tfrac{\|\Delta_{\text{adv}} - \Delta_{(1)}\| - \sigma_{\text{ref}}}
          {\|\Delta_{\text{adv}} - \Delta_{(1)}\|},\, 0,\, 0.75\right)$\;
  $\Delta_{\text{adv}} \leftarrow (1{-}\alpha)\Delta_{\text{adv}} 
                                  + \alpha \Delta_{(1)}$\;
}
$\|\Delta\|_{\text{target}} \leftarrow 
   \max(\|\Delta\|_{\text{med}},\, 0.5\|\Delta_{\text{loc}}\|)$\;
$\Delta_{\text{adv}} \leftarrow \Delta_{\text{adv}} \cdot 
   \|\Delta\|_{\text{target}} / \|\Delta_{\text{adv}}\|$\;
\Return $w_{\text{adv}} = w_t + \Delta_{\text{adv}}$\;

\caption{Selection-Aware Two-Stage Krum-Proxy Attack}
\label{alg:krum-proxy}
\end{algorithm}

\subsection{Problem Setup}
Let $w_t$ denote the global model in round $t$ and 
$\Delta_i = w_i - w_t$ the local update of client $i$. Krum selects the update minimizing the average squared distance to its $k$ nearest neighbors:
\begin{equation}
\text{score}(\Delta_i) = \frac{1}{k} \sum_{j \in \mathcal{N}_k(i)} 
  \|\Delta_i - \Delta_j\|^2
\end{equation}
Multi-Krum selects the $m$ updates with the lowest scores and averages them. The adversary seeks an update $\Delta_{\text{adv}}$ that preserves the effectiveness of the backdoor while minimizing this score to maximize the probability of selection.

\subsection{Stochastic Reference Construction}
\label{sec:refs}
The adversary constructs a reference set $\{\Delta_r\}_{r=1}^R$ by training $R$ clean models on disjoint partitions of its accessible data, starting from $w_t$. To capture benign heterogeneity, each reference uses an independently sampled learning rate 
$\eta_r \sim \mathcal{U}[\eta_{\min}, \eta_{\max}]$ and a local epoch count $E_r \sim \mathcal{U}\{E_{\min}, \dots, E_{\max}\}$. The set is reconstructed each round to track the evolving global model weights $w_t$. From it we derive the centroid $\bar{\Delta}_{\text{ref}}$, spread 
$\sigma_{\text{ref}} = \text{std}_r\|\Delta_r - \bar{\Delta}_{\text{ref}}\|$, 
and median norm $\|\Delta\|_{\text{med}} = \text{median}_r\|\Delta_r\|$, all of which are used in subsequent stages. The training anchor is constructed once as $\Delta_{\text{anchor}} = 0.85\Delta_{(1)} + 0.15\Delta_{(2)}$, where $\Delta_{(1)}, \Delta_{(2)}$ are the two references with the lowest Krum scores on the reference set itself. We use a blended anchor rather than strictly adhering to the single optimal reference $\Delta_{(1)}$. The $0.85$/$0.15$ blend keeps the anchor close to the lowest-score reference while using the second-lowest reference to smooth the gradient and reduce overfitting to an isolated trajectory.

\subsection{Krum-Proxy Objective}
True Krum scores require observing other clients' updates and are unavailable to the adversary. We approximate them using the reference set: for candidate $\Delta_{\text{adv}}$, we compute distances to all references and average the closest $\lfloor k/2 \rfloor$ among the nearest top-$k$ neighbors (Algorithm~\ref{alg:krum-proxy}, lines 14--15). The two-step structure focuses on optimizing the densest benign neighborhood while preserving differentiability through selection. Unlike MinMax/MinSum \cite{b16}, which treat neighbors uniformly, and the method of Fang et al. \cite{b5}, which requires observing other clients' updates and uses non-differentiable line search, our proxy operates from a local reference set and supports gradient-based joint optimization with the backdoor objective. Although the proxy objective explicitly minimizes the Krum distance, this provides a strong approximation of selection likelihood under Multi-Krum as well, since Multi-Krum relaxes the selection criteria by averaging the $m$ updates with the lowest Krum scores. Using only the inner half of the nearest neighbors emphasizes the densest reference region and the full-$k$ ablation in Section~\ref{ablationstudy} shows that averaging over more neighbors dilutes this signal and increases instability.

\subsection{Two-Stage Optimization}
\textbf{Stage 1 (Backdoor Injection)} trains for $E_1$ local epochs on mixed batches of clean and trigger-poisoned samples under cross-entropy loss alone with SGD ($\eta$, momentum $0.9$). Isolating the backdoor objective allows it to converge without competing gradient signals from the geometry losses, which have substantially smaller magnitudes early in training and would otherwise be suppressed (confirmed by ablation in Section~\ref{ablationstudy}).

\textbf{Stage 2 (Geometry Shaping)} trains for $E_2$ local epochs at half the learning rate under composite loss
\begin{equation}
\mathcal{L} = \lambda_l \mathcal{L}_{\text{CE}} + 
\lambda_n \mathcal{L}_{\text{norm}} + 
\lambda_k \mathcal{L}_{\text{krum}} + 
\lambda_a \mathcal{L}_{\text{anchor}},
\end{equation}
where 
$\mathcal{L}_{\text{norm}} = (\|\Delta_{\text{adv}}\| - 
\|\Delta_{\text{loc}}\|)^2$ matches the norm to a locally computed benign-style delta and $\mathcal{L}_{\text{anchor}} = \|\Delta_{\text{adv}} - \Delta_{\text{anchor}}\|^2$ pulls the update towards the dense core. $(\lambda_l, \lambda_n, \lambda_k, \lambda_a)$ are scalar hyperparameters that balance the contributions of the corresponding components in the objective function.

\subsection{Projection-Based Stealth Enforcement}
After Stage 2, a two-step projection (Algorithm~\ref{alg:krum-proxy}, lines 22--28) enforces stealth without erasing the backdoor. First, if the adversarial update deviates from the Krum-optimal reference $\Delta_{(1)}$ by more than $\sigma_{\text{ref}}$, it is interpolated toward $\Delta_{(1)}$ with the pull factor capped at $0.75$, keeping the update near the densest reference while preserving at least $25\%$ of the learned adversarial direction. The projection uses a single Krum-optimal reference rather than the 
blended training anchor because hard alignment at inference benefits from projecting to the densest single point, whereas the blend exists to provide a smoother gradient signal during training. Second, the result is scaled again to the median benign norm with a floor at $0.5\|\Delta_{\text{loc}}\|$, preventing collapse to trivially small updates that would erase the backdoor signal. 

\section{Experimental Setup}

\subsection{Federated Learning Architecture}

\begin{figure}[t]
    \centering
    \includegraphics[width=\linewidth]{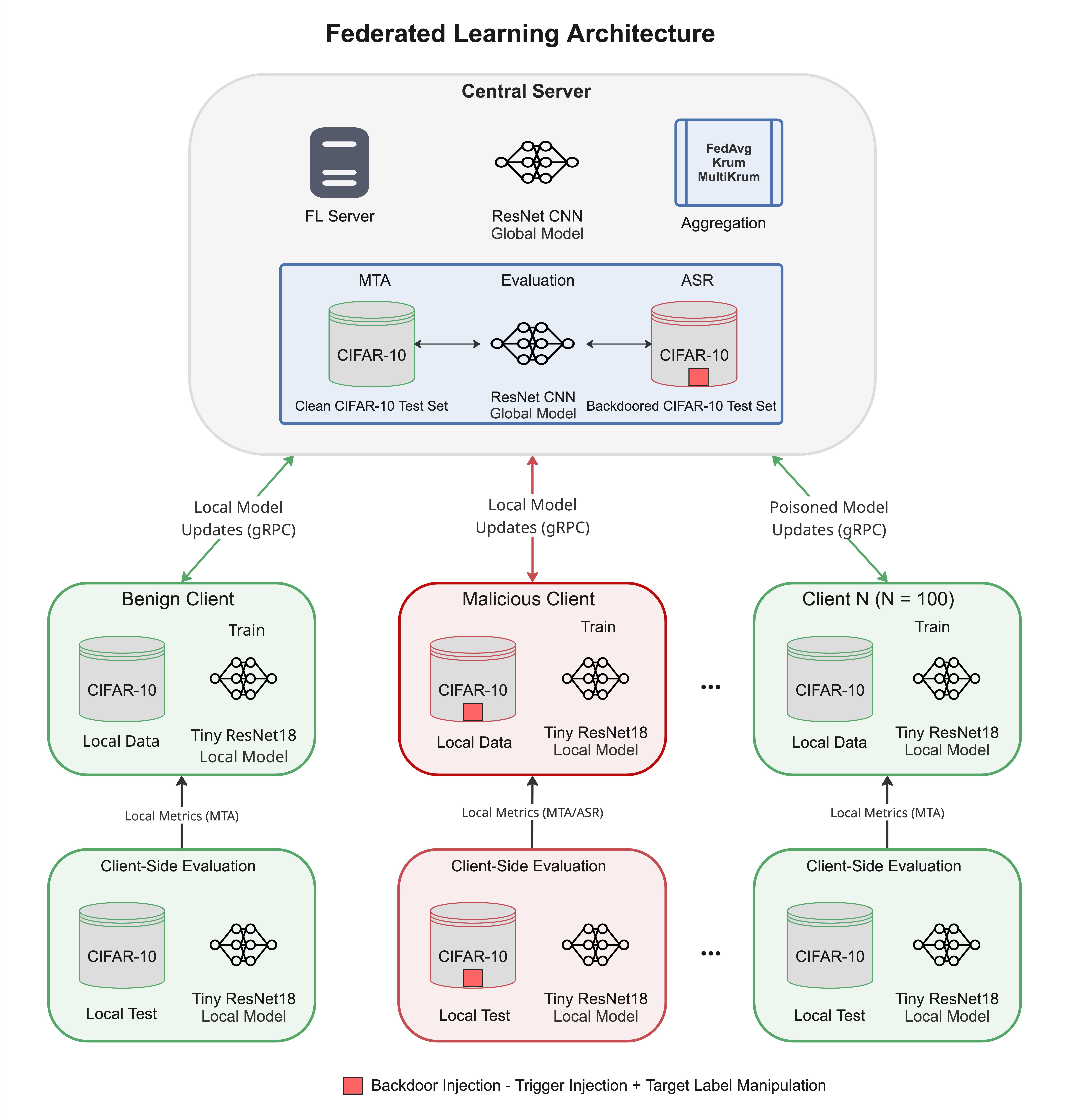}
    \caption{Overview of the experimental federated learning architecture.}
    \label{fig:fl_architecture}
\end{figure}

We simulate a federated learning setup comprising a central server and $N=100$ clients, as illustrated in Fig.~\ref{fig:fl_architecture}. In each round, a subset of 10 clients (sampling fraction $0.1$) is selected to perform local training. The server aggregates updates using Krum and Multi-Krum, where the number of Byzantine clients is set equal to the number of malicious clients per round, and $k = n - f - 2$ is used for neighbor selection. This ensures that the attack is evaluated under the maximum adversarial budget tolerated by Krum.

\subsection{Dataset, Partitioning and Model}

We evaluate the proposed attack on the CIFAR-10, MNIST, and EMNIST datasets to assess performance across varying data modalities and complexities. CIFAR-10 is used as the primary benchmark, while MNIST and EMNIST provide additional evaluation on grayscale digit and character recognition tasks. Each dataset is partitioned across clients using a Dirichlet distribution with concentration parameters ranging from $\alpha = 0.3$ to $0.9$ to simulate varying degrees of non-IID data heterogeneity. Each client performs an $80/20$ train-test split (seed = 42). For CIFAR-10, the training data is augmented using random cropping, horizontal flipping, and color jittering, while MNIST and EMNIST are used without augmentation.

We use a lightweight ResNet architecture (Tiny ResNet-18) with a base channel width of 8, resulting in approximately $0.27$M parameters. This reduced model size enables efficient simulation while maintaining sufficient clean accuracy for the CIFAR-10 classification.

\subsection{Client Training}

Experiments are conducted over 100 communication rounds. All methods are initialized from the same global model state at the beginning of the attack window. This state is obtained after the initial benign FL rounds and corresponds to the normal global model available to clients when the attack begins. All benign clients use SGD with momentum $0.9$, weight decay $5 \times 10^{-4}$, and label smoothing $0.05$. A cosine annealing learning rate scheduler is applied per training step for benign training. The batch size is fixed at 64. To model realistic client variability, learning rates and local epochs are randomly sampled. For Krum-Proxy, malicious clients use fixed learning rates and no label smoothing during attack optimization. All attacks use the same client sampling procedure, malicious-client budget, poison fraction, attack window, and attack-start global model. These hyperparameters were selected from preliminary tuning runs and then fixed across the reported Krum and Multi-Krum evaluations.

\subsection{Attack Configuration}

We use a trigger-based targeted attack with target label 2, poison fraction 10\%, 20 backdoor samples per batch, and 3 malicious clients per round, staying within Krum's assumed Byzantine budget.

We use $(\lambda_l, \lambda_n, \lambda_k, 
\lambda_a) = (0.4, 0.1, 0.25, 0.05)$ and $E_1 = E_2 = 1$, with 
gradient norm clipping at $\tau = 5.0$ in both stages. The same values are used across the reported Krum and Multi-Krum evaluations. The proposed attack is applied during local training using shared reference deltas ($R=6$) constructed from partitions of the malicious clients sampled. These references incorporate both inter-client data heterogeneity and stochastic training variations.

\subsection{Evaluation Metrics}

We centrally evaluated the global model after each communication round using clean model test accuracy (MTA) in the CIFAR-10 test samples and attack success rate (ASR) in trigger-poisoned CIFAR-10 test samples, as shown in Fig.~\ref{fig:fl_architecture}.

\subsection{Baselines}

We evaluate FedAvg, Krum, and Multi-Krum aggregation, and compare the proposed attack with two poisoning baselines: Scaled Backdoor, which uniformly scales poisoned updates, and Constrain-and-Scale \cite{b3}, which enforces norm and direction constraints for stealth.

\section{Results and Analysis}

\subsection{Attack Effectiveness}

\begin{figure}[t]
    \centering
    \includegraphics[width=\linewidth]{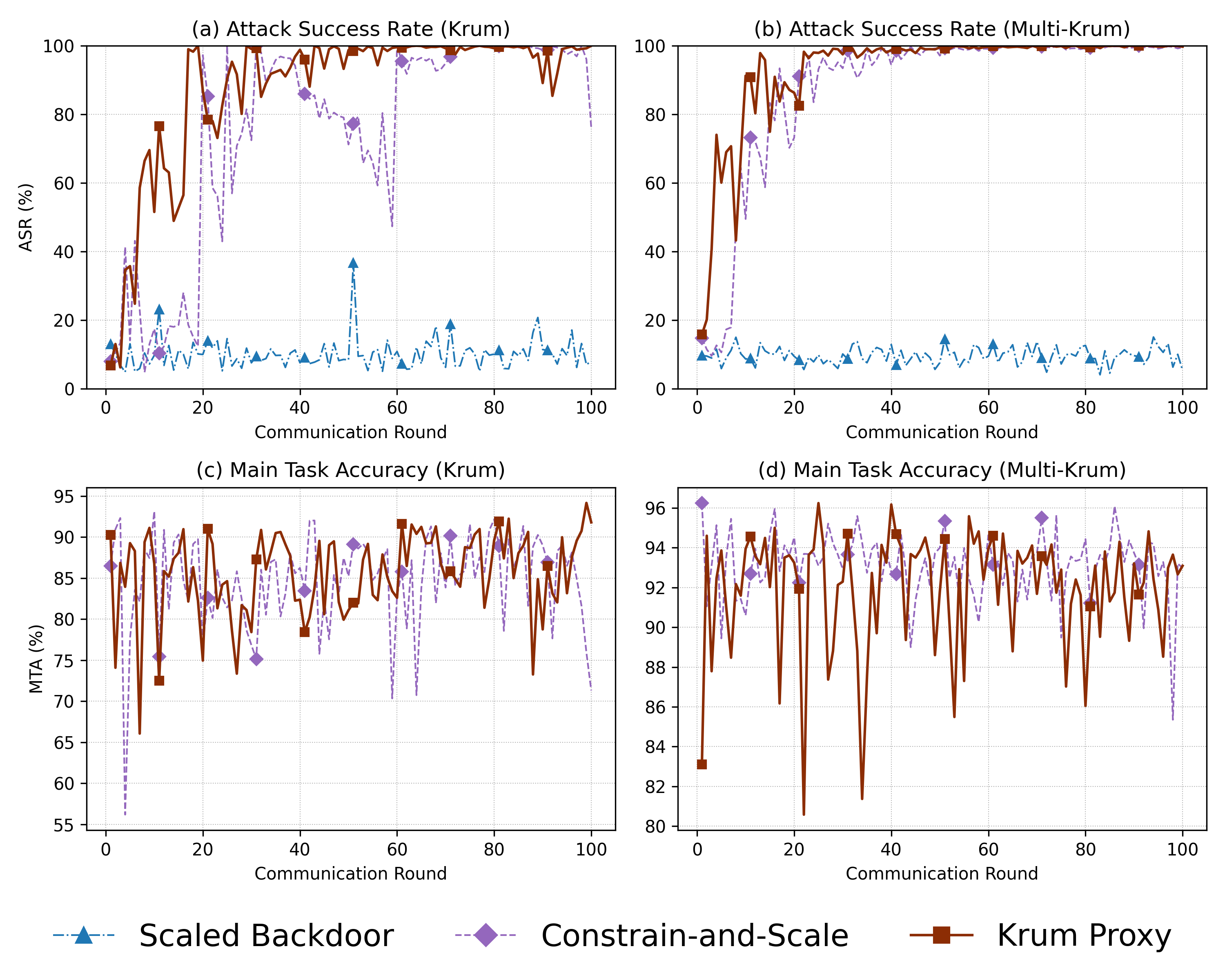}
\caption{Attack performance under robust aggregation. 
(a) Attack success rate under Krum. 
(b) Attack success rate under Multi-Krum. 
(c) Main task accuracy under Krum. 
(d) Main task accuracy under Multi-Krum.
Legends indicate different attack methods (Scaled Backdoor, Constrain-and-Scale, and Krum-Proxy)}
    \label{fig:attack_performance}
\end{figure}

Figures~\ref{fig:attack_performance}(a)-(d) show ASR and MTA throughout training under Krum and Multi-Krum. The Scaled Backdoor baseline remains largely ineffective under robust aggregation, while Constrain-and-Scale achieves only partial and unstable evasion, particularly under Krum. In contrast, the proposed Krum-Proxy attack rapidly converges to high ASR under both defenses while maintaining competitive MTA throughout training, demonstrating that directly optimizing the aggregation selection objective enables consistent evasion without sacrificing model utility. Table~\ref{tab:persistence} summarizes convergence and post-convergence stability across methods and shows that Krum-Proxy improves injection stability, enabling persistent attack success once selected rather than relying on intermittent reinjection.

\begin{table}[htbp]
\caption{Final Performance Summary After 100 Communication Rounds}
\centering
\footnotesize
\setlength{\tabcolsep}{4pt}
\begin{tabular}{|c|p{2.3cm}|c|c|}
\hline
\textbf{Aggregation} & \textbf{Attack} & \textbf{MTA (\%)} & \textbf{ASR (\%)} \\
\hline
FedAvg & Scaled Backdoor & 92.17 & 98.52 \\
FedAvg & Constrain \& Scale & 92.73 & 100 \\
FedAvg & \textbf{Krum-Proxy} & 94.43 & 95.64 \\
\hline
Krum & Scaled Backdoor & 84.00 & 7.91 \\
Krum & Constrain-and-Scale & 71.31 & 75.99 \\
Krum & \textbf{Krum-Proxy} & 91.79 & 100 \\
\hline
Multi-Krum & Scaled Backdoor & 94.44 & 5.35 \\
Multi-Krum & Constrain-and-Scale & 92.99 & 99.54 \\
Multi-Krum & \textbf{Krum-Proxy} & 93.09 & 99.74 \\
\hline
\multicolumn{4}{l}{\footnotesize MTA: Main Task Accuracy; ASR: Attack Success Rate.}
\end{tabular}
\label{tab:final_results}
\end{table}

\begin{table}[t]
\centering
\caption{Post-Convergence Persistence (Krum)}
\label{tab:persistence}
\footnotesize
\setlength{\tabcolsep}{2pt}
\begin{tabular}{|c|c|c|c|}
\hline
\textbf{Method} & \textbf{ASR (\%)} & \textbf{Rounds$_{ASR}$} & \textbf{Rounds$_{MTA}$} \\
\hline
Constrain-and-Scale & 92.2 $\pm$ 13.1 & 24 & 11 \\
\hline
Krum-Proxy  & 98.6 $\pm$ 2.8  & 63 & 6 \\
\hline
\multicolumn{4}{l}{\footnotesize Rounds$_{ASR}$: Number of rounds where ASR $>$ 90\%.}\\
\multicolumn{4}{l}{\footnotesize Rounds$_{MTA}$: Number of rounds where MTA drops $<$ 75\%.}
\end{tabular}
\end{table}

\subsection{Ablation Study}
\label{ablationstudy} 

Table~\ref{tab:ablation} reports post-convergence MTA stability, measured as the number of rounds with MTA below 75 after the attack first reaches ASR $\geq 90$, together with the average MTA and ASR across 100 communication rounds under Krum aggregation. The results show that removing key components of our proposed method substantially degrades the optimization stability and attack reliability.

\begin{table}[t]
\caption{Ablation analysis of Krum-Proxy Components under Krum aggregation.}
\centering
\footnotesize
\setlength{\tabcolsep}{4pt}
\begin{tabular}{|p{3.2cm}|c|c|c|}
\hline
\textbf{Variant} & \textbf{Avg. MTA} & \textbf{Avg. ASR} & \textbf{MTA Crash}\\
\hline
\textbf{Krum-Proxy (Full)} & 85.61 & 88.07 & 3\\
\hline
w/o Two-Stage (joint loss) & 78.73 & 84.87 & 22\\
\hline
w/o $\mathcal{L}_{\text{krum}}$  & 80.59 & 86.17 & 17\\
\hline
w/o $\mathcal{L}_{\text{anchor}}$ + Projection & 86.39 & 81.60 & 3\\
\hline
Full-$k$-Proxy ($k$ neighbors)& 82.43 & 82.10 & 12\\
\hline
w/o Stochastic References& 81.76 & 76.12 & 14 \\
\hline
\multicolumn{4}{l}{\footnotesize Avg. MTA: Mean main task accuracy across all rounds;}\\
\multicolumn{4}{l}{\footnotesize Avg. ASR: Mean attack success rate across all rounds;}\\
\multicolumn{4}{l}{\footnotesize MTA Crash: Post-convergence rounds with MTA $<75 \%$.}
\end{tabular}
\label{tab:ablation}
\end{table}

Eliminating the two-stage optimization causes the most severe instability, producing 22 post-convergence MTA crashes and reducing average MTA by 6.88 percentage points. This indicates that joint optimization disrupts early backdoor convergence due to competing gradients. Removing $\mathcal{L}_{\text{krum}}$ produces 17 post-convergence crashes and lowers average ASR from 88.07\% to 86.17\%, indicating that without explicit proxy minimization the attack relies more heavily on projection and becomes less stable across rounds.

Replacing stochastic reference construction with a single deterministic reference causes the largest ASR degradation overall, reducing average ASR to 76.12\%, delaying convergence until round 27, and producing 14 post-convergence crashes. This indicates that stochastic references are necessary to capture benign variability. Using a full-$k$-proxy instead of the proposed inner half-$k$ averaging results in 12 post-convergence crashes and reduces round-100 MTA to 68.75\%, demonstrating that averaging over all neighbors dilutes the optimization signal.

Finally, removing anchor alignment and projection together yields the highest average MTA (86.39\%) and only 3 post-convergence crashes, but reduces average ASR by 6.47 percentage points and delays convergence by four rounds, indicating that these mechanisms primarily improve early-round selection consistency rather than late-round model stability.

Under an MTA crash threshold of $< 60\%$, the proposed full method exhibits zero catastrophic post-convergence failures, while the single-stage and no-$\mathcal{L}_{\text{krum}}$ variants produce 10 and 7 such crashes, respectively. For example, the single-stage variant reaches 90\% ASR but later falls to 27\% MTA, showing that backdoor success can result in unusable clean-task performance.

\subsection{Heterogeneity Sensitivity}
The results demonstrate that the proposed attack remains highly effective across varying levels of data heterogeneity as well. Under mild heterogeneity ($\alpha=0.7$), the model achieves 90.02\% MTA with 99.94\% ASR. As heterogeneity increases to $\alpha=0.5$, MTA slightly improves to 90.68\% while ASR remains high at 98.02\%. Under strong heterogeneity ($\alpha=0.3$), MTA decreases to 87.78\%, but the attack continues to achieve a high ASR of 98.20\%, demonstrating robustness to non-IID data distributions.

\subsection{Cross-Dataset Generalization}
To verify that the proposed attack is not specific to CIFAR-10, we also evaluated it on MNIST and EMNIST in the same Krum-based federated setting. On MNIST, the attack similarly achieves rapid convergence, surpassing 90\% ASR by round 18 and maintaining near-perfect attack success thereafter while preserving high clean accuracy (98.99\% final MTA, 99.95\% final ASR at round 100). On EMNIST, which introduces greater class diversity and heterogeneity, the attack rapidly exceeds 90\% ASR by Round 7 and maintains sustained high ASR ($>98\%$) while preserving stable model accuracy (approximately 87 to 90\% MTA). These results demonstrate that our method remains effective even in more complex and heterogeneous data distributions.

\subsection{Generalization and Aggregation Evasion}

Krum-Proxy improves aggregation-level evasion. Under Krum, malicious selection increases from 9 to 29 rounds. Under Multi-Krum, malicious inclusion increases from 62/300 to 175/300 updates, raising the inclusion rate from 20.7\% to 58.3\%. 

To evaluate whether the attack transfers beyond Krum-style aggregation, we also test the Trimmed Mean aggregation. Krum-Proxy remains effective, reaching 97.65\% final ASR with 94.99\% MTA and no MTA crashes below 75\%. 
 
To evaluate sensitivity to the malicious-client budget, we vary the number of malicious clients under Krum and Multi-Krum. Under Krum, $1$, $2$, and $3$ malicious clients per round reach 98.17\%, 98.21\%, and 100\% final ASR while maintaining 91.79\%, 86.29\%, and 91.79\% final MTA, respectively. Under Multi-Krum, the corresponding runs reach 99.21\%, 99.84\%, and 99.74\% final ASR while maintaining 89.80\%, 93.96\%, and 93.09\% final MTA.

\section{Discussions and Limitations} 

The proposed attack has some practical limitations. Its effectiveness depends on the quality of locally constructed clean reference updates. Under strong non-IID settings or high client heterogeneity, these estimates may deviate from the true global distribution, reducing selection probability. The attack is also sensitive to tightly clustered benign updates, where even small deviations can prevent selection despite maintaining overall stealth. There is also a trade-off between stealth and backdoor strength, as stronger geometric constraints improve selection but may weaken the attack signal.

The proposed method is designed for Krum-style distance-based selection and does not explicitly optimize coordinate-wise rules such as Median, Trimmed Mean, or Bulyan.

The proposed attack also introduces additional computation due to reference construction and proxy-based optimization and relies on local approximations of server-side behavior, which may not fully capture the true aggregation dynamics. Krum-Proxy required $1.29\times$ the end-to-end runtime of Constrain-and-Scale in the CIFAR-10 Krum setting, reflecting the cost of reference construction and proxy-based refinement.

These observations suggest that future defenses should incorporate richer signals beyond single-round distance metrics, such as temporal consistency or structural analysis of updates, to better detect defense-aware attacks.

\section{Conclusion}

This paper presents the Krum-Proxy attack, a selection-aware model poisoning strategy that bypasses Krum-based robust aggregation by explicitly optimizing adversarial updates for favorable geometric placement under the aggregation rule. Experimental results show that Krum-Proxy substantially improves attack success and stability under Krum, while increasing malicious inclusion under Multi-Krum and preserving competitive model utility. These findings demonstrate that distance-based robust aggregation remains vulnerable to adaptive adversaries that directly optimize the defender's selection criterion. Future work should explore defense mechanisms that incorporate richer signals beyond single-round distance comparisons, such as temporal consistency or structural analysis of updates, to detect such attacks efficiently.

\vspace{12pt}
\end{document}